\documentclass[12pt,a4paper]{article}

\usepackage[british]{babel}
\usepackage{rotating}
\usepackage{adjustbox}

\usepackage[a4paper,top=2cm,bottom=2cm,left=2.5cm,right=2.5cm,marginparwidth=1.75cm]{geometry}

\usepackage[style=ieee, backend=biber, maxcitenames=1, mincitenames=1, maxbibnames=999, minbibnames=1]{biblatex}
\DeclareFieldFormat[article, inbook, incollection, inproceedings, misc, thesis, unpublished]{title}{#1}  
\DeclareFieldFormat{title}{#1}  
\DeclareFieldFormat{journaltitle}{#1}  
\DeclareFieldFormat{booktitle}{#1}  
\DeclareFieldFormat{maintitle}{#1}  
\DeclareFieldFormat{issuetitle}{#1}  
\DeclareFieldFormat{eventtitle}{#1}  

\DeclareFieldFormat{url}{\url{#1}}
\DeclareFieldFormat{doi}{\newline\mkbibacro{DOI}\addcolon\space\url{#1}}

\usepackage[table,xcdraw]{xcolor}
\usepackage{float}
\usepackage{makecell}
\usepackage{amsmath}
\usepackage{graphicx}
\usepackage[colorlinks=true, allcolors=blue]{hyperref}
\usepackage{hyperref}
\usepackage{orcidlink}
\usepackage[title]{appendix}
\usepackage{mathrsfs}
\usepackage{amsfonts}
\usepackage{booktabs} 
\usepackage{caption}  
\usepackage{threeparttable} 
\usepackage{algorithm}
\usepackage{algorithmicx}
\usepackage{algpseudocode}
\usepackage{listings}
\usepackage{enumitem}
\usepackage{chngcntr}
\usepackage{booktabs}
\usepackage{lipsum}
\usepackage{subcaption}
\usepackage{authblk}
\usepackage[T1]{fontenc}    
\usepackage{csquotes}       
\usepackage{diagbox}
\usepackage{tabularx}
\usepackage{multirow}
\usepackage{rotating}
\usepackage{bm}
\usepackage{wrapfig}

\usepackage{setspace}
\usepackage{titlesec}
\titleformat{\section} 
  {\normalfont\Large\bfseries}{\thesection.}{1em}{}
  
\usepackage{lineno} 

\rightlinenumbers 

\linenumbers 

\usepackage{float}   
\usepackage{caption} 


\newcolumntype{C}[1]{>{\centering\arraybackslash}m{#1}}

\title{ARCO-BO: Adaptive Resource-aware COllaborative Bayesian Optimization for Heterogeneous Multi-Agent Design}

\author[1]{Zihan Wang}
\author[1]{Yi-Ping Chen}
\author[1]{Tuba Dolar}
\author[1,*]{Wei Chen}
\affil[1]{\small Department of Mechanical Engineering, Northwestern University,
   Evanston, IL, 60208}
\affil[*]{Corresponding author}

\date{}  

\begin{document}
\maketitle
\begin{abstract}
Modern scientific and engineering design increasingly involves distributed optimization, where agents such as laboratories, simulations, or industrial partners pursue related goals under differing conditions. These agents often face heterogeneities in objectives, evaluation budgets, and accessible design variables, which complicates coordination and can lead to redundancy, poor resource use, and ineffective information sharing.
Bayesian Optimization (BO) is a widely used decision-making framework for expensive black box functions, but its single-agent formulation assumes centralized control and full data sharing. Recent collaborative BO methods relax these assumptions, yet they often require uniform resources, fully shared input spaces, and fixed task alignment, conditions rarely satisfied in practice.
To address these challenges, we introduce Adaptive Resource Aware Collaborative Bayesian Optimization (ARCO-BO), a framework that explicitly accounts for heterogeneity in multi-agent optimization. ARCO-BO combines three components: a similarity and optima-aware consensus mechanism for adaptive information sharing, a budget-aware asynchronous sampling strategy for resource coordination, and a partial input space sharing for heterogeneous design spaces. Experiments on synthetic and high-dimensional engineering problems show that ARCO-BO consistently outperforms independent BO and existing collaborative BO via consensus approach, achieving robust and efficient performance in complex multi-agent settings.
\end{abstract}


\section{Introduction}

\begin{figure}[H]
\centering
\includegraphics[width=1\linewidth]{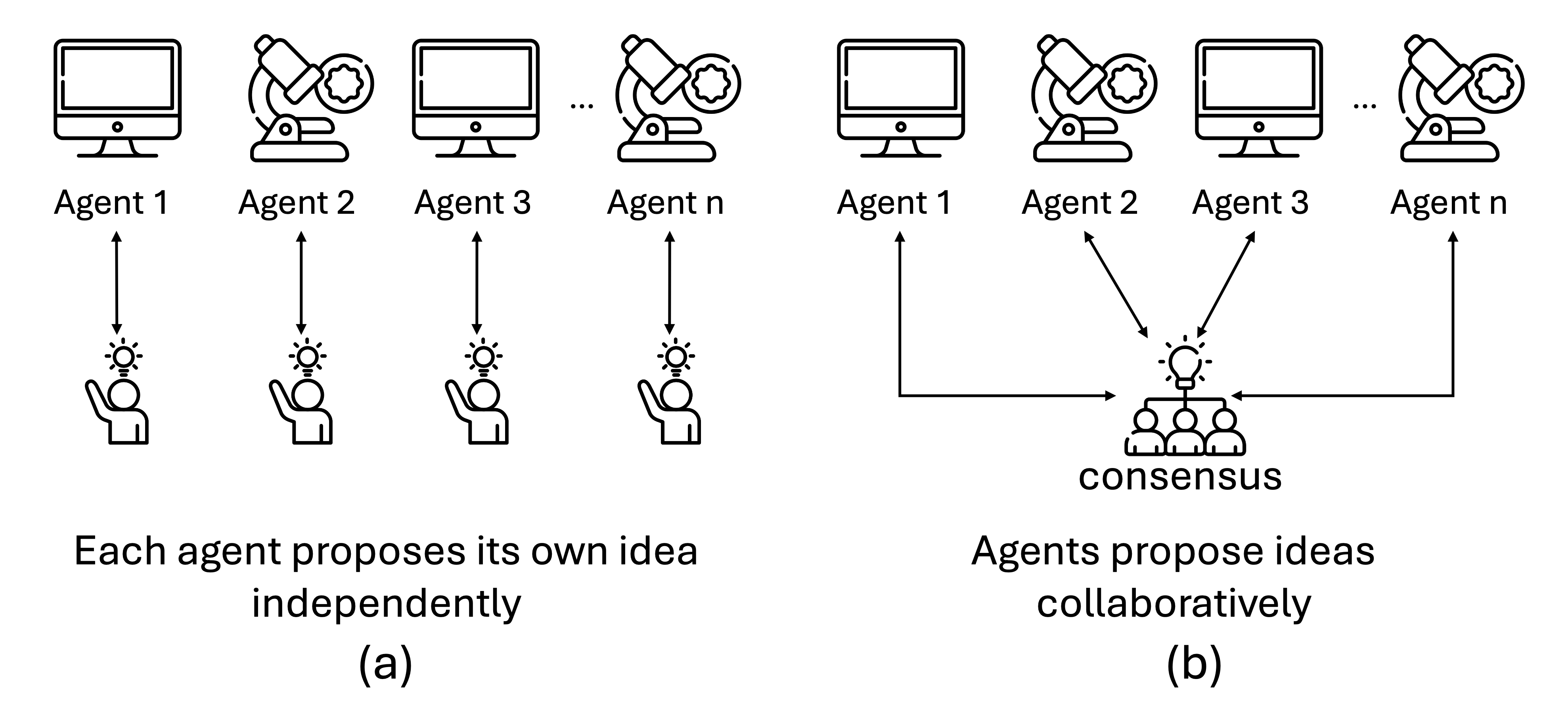}
\caption{Conceptual illustration of multi-agent system. {\normalfont (a) Without collaboration, agents operate independently. (b) With collaboration, agents share information through consensus.}}
\label{fig:concept_collaborative}
\end{figure}

Modern scientific and engineering challenges increasingly demand coordinated efforts across multiple teams, disciplines, or organizations. Such collaboration is common in domains like multi-fidelity design optimization \cite{kvan2000collaborative, chen2024latent}, materials discovery \cite{noack2021autonomous,khatamsaz2025towards,coley2020autonomous}, and distributed manufacturing \cite{kontar2021internet, liu2024scalable, yue2025collaborative}. In multi-fidelity design, for instance, low-cost simulations provide broad exploration via faster/cheaper design evaluation, while high-fidelity experiments target promising candidates by focusing more on exploitation. In materials discovery, multiple laboratories may conduct experiments under different conditions or with distinct tools, each producing partial insights into a shared design space. In distributed manufacturing, facilities specialize in different stages of a product’s life cycle, each facing unique constraints and goals. 
These settings are a natural fit for a multi-agent design perspective \cite{juziuk2014design,deloach2005multi}, where design agents (e.g., teams, labs, or models) contribute complementary, potentially heterogeneous information toward a common design objective.
Heterogeneity can take many forms, including \textit{function heterogeneity} when agents pursue related but distinct objectives, \textit{budget heterogeneity} when resources are unequal, and \textit{input-space heterogeneity} when exploration occurs in overlapping but not fully shared domains.
In this work, we focus on collaboration rather than competition: agents coordinate to improve collective performance despite these differences.
Figure~\ref{fig:concept_collaborative} contrasts independent operation, where agents act in isolation, with collaborative operation, where agents exchange information through a consensus mechanism.

A variety of approaches have been developed for multi-agent design \cite{yang2019survey,chang2014multi,nedic2009distributed,amirkhani2022consensus,qin2016recent,li2019survey,olfati2007consensus,semsar2009multi,boella2007game,niazi2011agent,cardoso2021review}. Distributed optimization \cite{yang2019survey,chang2014multi,nedic2009distributed} enables agents to solve a global problem via local updates and limited communication; methods such as distributed gradient descent and the Alternating Direction Method of Multipliers (ADMM) are widely used. Consensus algorithms \cite{amirkhani2022consensus,qin2016recent,li2019survey,olfati2007consensus} align shared design variables through local interactions, often without centralized control. Game-theoretic frameworks \cite{semsar2009multi,boella2007game} model strategic agents and study equilibria and incentive-compatible mechanisms when cooperation cannot be assumed. Agent-based modeling \cite{niazi2011agent,cardoso2021review} provides a flexible simulation framework for heterogeneous agents governed by decentralized decision rules.
While effective in many settings, these methods rely on access to objectives or gradients and are not well suited for costly or heterogeneous tasks.
Recently, multi-agent reinforcement learning (MARL) \cite{canese2021multi,bucsoniu2010multi,zhang2021multi} has emerged as a data-driven approach that learns coordination directly from interaction, without explicit models of the environment or other agents. It suits dynamic or partially observable settings where handcrafted rules are impractical. However, many MARL methods emphasize reward maximization and fall short in sample efficiency and calibrated uncertainty, limiting their effectiveness in expensive design tasks.
In practice, agents often face heterogeneous conditions. Coordinating these distributed efforts to achieve collective progress while minimizing redundancy is difficult, and exhaustive search or simple heuristics are often inefficient or brittle.
In contrast, adaptive and resource-aware optimization approaches \cite{gligorea2023adaptive,martin2020systematic}, which guide evaluations based on feedback, uncertainty, and budget constraints, offer a promising path toward efficient and scalable multi-agent design.

Bayesian Optimization (BO) has been a popular effective adaptive learning framework for data-efficient decision-making in scenarios where evaluating objective functions is expensive \cite{frazier2018tutorial,pelikan2005hierarchical,frazier2018bayesian}. BO leverages probabilistic surrogate models such as Gaussian Processes (GPs) \cite{seeger2004gaussian} to estimate uncertainty and guides exploration through acquisition functions that balance exploration and exploitation to efficiently locate global optima with minimal samples. However, standard BO frameworks inherently assume a centralized, single-agent setup in which all evaluations and optimization decisions are controlled by a unified model operating over the entire design space. This assumption limits the applicability of traditional BO in distributed and collaborative environments where optimization tasks are performed by multiple agents or in a parallelized setup.

To overcome this limitation, collaborative BO \cite{yue2025collaborative,chen2025multi,zhan2025collaborative} introduces a new paradigm for multi-agent design by enabling agents to share surrogate information and coordinate sampling decisions \textit{without requiring centralized control or full data sharing}. The collaborative BO strategy proposed by \cite{chen2025multi} introduces constrained Gaussian process surrogates that enable agents to leverage informative evaluations discovered by high-performing collaborators. Rather than treating each agent's optimization process in isolation, this method facilitates cross-agent knowledge sharing by selectively incorporating promising designs observed by others, subject to compatibility constraints. Another collaboration strategy proposed by \cite{zhan2025collaborative} adopts a federated optimization paradigm, each agent trains its own model using its local data and then shares only certain parameters (like gradients or parts of the model) with others. By doing this, the agents work together to build a shared global model without directly sharing their raw data. More recently, consensus-based collaborative BO methods have been proposed \cite{yue2025collaborative} in which agents agree on their next-to-sample designs by sharing their local designs, but not design performances.

In parallel, emerging multi-source data fusion methods \cite{ravi2025interpretable} offer an alternative paradigm for knowledge sharing through unified surrogate models. A representative example is the Latent Variable Gaussian Process (LVGP) \cite{zhang2020latent}, which embeds discrete or high-dimensional variables into a continuous latent space to capture shared structures across heterogeneous tasks. In multi-fidelity optimization, LVGP has demonstrated strong potential for capturing shared structures and correlations across tasks or fidelity levels by implicitly learning task similarities through latent embeddings \cite{chen2024latent}. This enables effective information transfer across heterogeneous sources \cite{comlek2025heterogeneous}, thereby improving sample efficiency.
Nevertheless, unlike collaborative BO frameworks that safeguard agent-level privacy by avoiding the exchange of true evaluation outcomes, LVGP requires access to both the input variables and the observed outputs from all agents in order to train a unified surrogate model. Additionally, the joint optimization of latent embeddings and hyperparameters may become computationally intensive as the number of agents increases, limiting LVGP’s scalability and applicability in many-agent systems.

While these advances mark notable progress in collaborative optimization, extending them to real-world multi-agent scenarios continues to pose substantial challenges. In particular, strong function heterogeneity across agents remains a hurdle for resource distribution in active learning under collaborative setup \cite{al2024collaborative}. Beyond differing objectives, collaborative optimization often faces other forms of heterogeneity that further complicate coordination. In materials discovery, for instance, laboratories may pursue aligned optimization goals but operate under distinct process constraints, experimental budgets, and available input variables due to differences in equipment or protocols \cite{talapatra2018autonomous,lookman2019active}. In distributed manufacturing, concerns around data privacy and intellectual property frequently restrict the extent to which performance data or design specifications can be shared across organizational boundaries \cite{smith2017federated,kusne2023scalable, zemskov2024security}. Multi-fidelity design presents yet another challenge, where computational agents can explore broad regions of the design space, while physical evaluations are limited to a smaller, costly subset \cite{schmidt2019recent}. These scenarios highlight three recurring and interdependent forms of heterogeneity that collaborative Bayesian optimization must confront: (1) heterogeneity in objective functions, (2) heterogeneity in evaluation budgets, and (3) heterogeneity in shareable input variables. However, existing work has mainly focused on limited function heterogeneity or privacy-aware coordination, without simultaneously addressing all three aspects in a unified framework.

In this paper, we propose Adaptive Resource-Aware Collaborative Bayesian Optimization (ARCO-BO), a framework specifically designed for multi-agent optimization in heterogeneous settings to handle the aforementioned limitations. As existing collaborative BO methods typically assume shared objectives, uniform resource budgets, or fully overlapping input spaces, ARCO-BO explicitly tackles three critical forms of heterogeneity: differing objective functions, asymmetric evaluation budgets, and partially shared input domains. To address these challenges, ARCO-BO introduces three key capabilities. First, it employs an adaptive, similarity-aware consensus mechanism that adjusts the degree of information sharing based on the alignment of surrogate models, enabling dynamic and selective collaboration among agents. Second, it implements a resource-aware sampling strategy that schedules agent participation based on their available evaluation budgets, promoting efficient and balanced exploration. Third, it enables collaboration over partially shared input spaces, allowing agents to coordinate on common variables while independently optimizing their private inputs. We evaluate ARCO-BO on both low-dimensional illustrative examples and high-dimensional engineering problems, comparing its performance to two baselines: (1) standard Bayesian optimization with independent agents, and (2) conventional consensus-based collaborative BO with static update rules \cite{yue2025collaborative}. Our results show that ARCO-BO consistently achieves superior design quality and faster convergence across all tested scenarios.

The contributions of this work are summarized as follows:
\begin{itemize}
    \item We introduce ARCO-BO, a novel framework for multi-agent Bayesian optimization under heterogeneous settings, including function heterogeneity, asymmetric evaluation budgets, and partially shared input spaces.
    \item We develop three key algorithmic components: a similarity-aware consensus mechanism for dynamic information sharing, a budget-aware sampling strategy to coordinate evaluations based on remaining agent resources, and a partially shared input model to handle heterogeneous design spaces across agents.
    \item We demonstrate that ARCO-BO consistently outperforms both independent optimization and static-consensus baselines on synthetic benchmarks and high-dimensional engineering design problems.
\end{itemize}

The remainder of this paper is organized as follows. Section \ref{sec:technical_background} reviews the technical background, including the existing Bayesian optimization and collaborative Bayesian optimization via consensus. Section \ref{sec:ARCO-BO} introduces the proposed ARCO-BO framework, detailing its adaptive consensus mechanism, resource-aware sampling strategy, and support for partially shared input spaces. Section \ref{sec:exp} outlines the setup of numerical studies, including evaluation metrics and computational configurations. Section \ref{sec:case_study} presents and analyzes the results, comparing ARCO-BO to baseline methods across diverse multi-agent optimization scenarios. Section \ref{sec:conclusion} concludes the paper and discusses potential directions for future research.

\section{Technical Background}
\label{sec:technical_background}

This section introduces the foundational concepts that motivate and inform our proposed ARCO-BO. We begin with a review of BO, followed by its extension to collaborative multi-agent settings via consensus-based Collaborative BO. These components highlight the heterogeneity and coordination challenges that ARCO-BO is designed to address.

\subsection{Bayesian Optimization}

BO ~\cite{shahriari2015taking, frazier2018tutorial, mockus2005bayesian, movckus1975bayesian} 
has emerged as a powerful framework for the global optimization of expensive, black-box functions. 
In these settings, the objective function \( f \) is not analytically accessible and can only be evaluated 
through potentially noisy observations of the form \( y = f(\mathbf{x}) + \epsilon \), 
where \( \epsilon \sim \mathcal{N}(0, \sigma^2) \) denotes independent Gaussian noise. 
The goal is to identify the global minimizer of the function:
\begin{equation}
\mathbf{x}^* = \arg\min_{\mathbf{x}} f(\mathbf{x}),
\end{equation}
with \( \mathbf{x} \) residing in a \( d \)-dimensional design space. 
A central challenge in BO lies in the high cost of function evaluations, 
which severely limits the number of allowable queries. 
As a result, exhaustive search strategies become impractical, 
and conventional optimization methods often prove inefficient. BO addresses this challenge by leveraging a probabilistic surrogate model to approximate the unknown objective function. This model guides the selection of future query points in a way that strategically balances exploration of uncertain regions with exploitation of areas likely to yield high performance. BO relies on two fundamental components: a surrogate model (typically a Gaussian Process) and an acquisition function that quantifies the utility of sampling each candidate point.

The Gaussian Process (GP) defines a distribution over functions and is characterized by a mean function $\mu(\cdot)$ and a covariance function $k(\cdot, \cdot)$:
\begin{equation}
    f(\mathbf{x}_{1:k}) \sim \mathcal{GP}\left(\mu(\mathbf{x}), k(\mathbf{x}, \mathbf{x}')\right).
\end{equation}
Given an observed dataset $\mathcal{D} = \{(\mathbf{x}_i, y_i)\}_{i=1}^n$, the posterior distribution of $f$ at a set of new points $\mathbf{x}_{1:k}$ is:
\begin{equation}
    f(\mathbf{x}_{1:k}) \mid \mathcal{D} \sim \mathcal{N}\left( \mu_n(\mathbf{x}_{1:k}), \Sigma_n(\mathbf{x}_{1:k}, \mathbf{x}_{1:k}) \right),
\end{equation}
where $\mu_n$ and $\Sigma_n$ denote the posterior mean and covariance functions, respectively.

An acquisition function is a heuristic used to select the next sampling point. It takes the GP posterior as input and balances exploration and exploitation. Common acquisition functions \cite{frazier2018tutorial,frazier2018bayesian,snoek2012practical} include Probability of Improvement (PI), Expected Improvement (EI), Lower Confidence Bound (LCB), and Thompson Sampling (TS). We adopt EI throughout this study due to its sensitivity to potential improvements and its efficiency in converging to the optimum.

Assuming we have a minimization problem, the improvement utility function is:
\begin{equation}
    u(\mathbf{x}) = \max(0, f^* - f(\mathbf{x})),
\end{equation}
where $f^*$ is the incumbent, i.e., the best (minimum) observed value so far. EI is formulated as:
\begin{align}
    \alpha_{\text{EI}}(\mathbf{x}) &= \mathbb{E}[u(\mathbf{x}) \mid \mathcal{D}] \nonumber \\
    &= (f^* - \mu(\mathbf{x})) \Phi\left( \frac{f^* - \mu(\mathbf{x})}{\sigma(\mathbf{x})} \right) + \sigma(\mathbf{x}) \phi\left( \frac{f^* - \mu(\mathbf{x})}{\sigma(\mathbf{x})} \right),
\end{align}
where $\Phi(\cdot)$ and $\phi(\cdot)$ are the CDF and PDF of the standard normal distribution, and $\mu(\mathbf{x}), \sigma^2(\mathbf{x})$ are the GP posterior mean and variance. The next sampling point is chosen by maximizing $\alpha_{\text{EI}}(\mathbf{x})$.

\subsection{Collaborative Bayesian Optimization via Consensus}

As conventional BO is typically formulated for a single-agent setting, collaborative BO \cite{yue2025collaborative,zhan2025collaborative,chen2025multi} generalizes the BO framework to accommodate these multi-agent scenarios by allowing agents to share information of the query samples to improve convergence efficiency.

In the consensus-based CBO framework proposed by Yue \emph{et al}. \cite{yue2025collaborative}, each agent \( k \in \{1, \dots, K\} \) maintains its own surrogate model and acquisition function \( U_k(\mathbf{x}) \), selecting its candidate design point at iteration \( t \) as:

\begin{equation}
    \mathbf{x}_k^{(t)}
    =
    \arg\max_{\mathbf{x}}
    \mathbb{E}_{\hat{f}_k | D_k^{(t)}} \left[ U_k(\hat{f}_k(\mathbf{x})) \right],
\end{equation}
where \( \hat{f}_k \) is the surrogate model fitted on agent \( k \)'s dataset \( D_k^{(t)} \). 

\begin{figure}[H]
    \centering
    \includegraphics[width=0.9\linewidth]{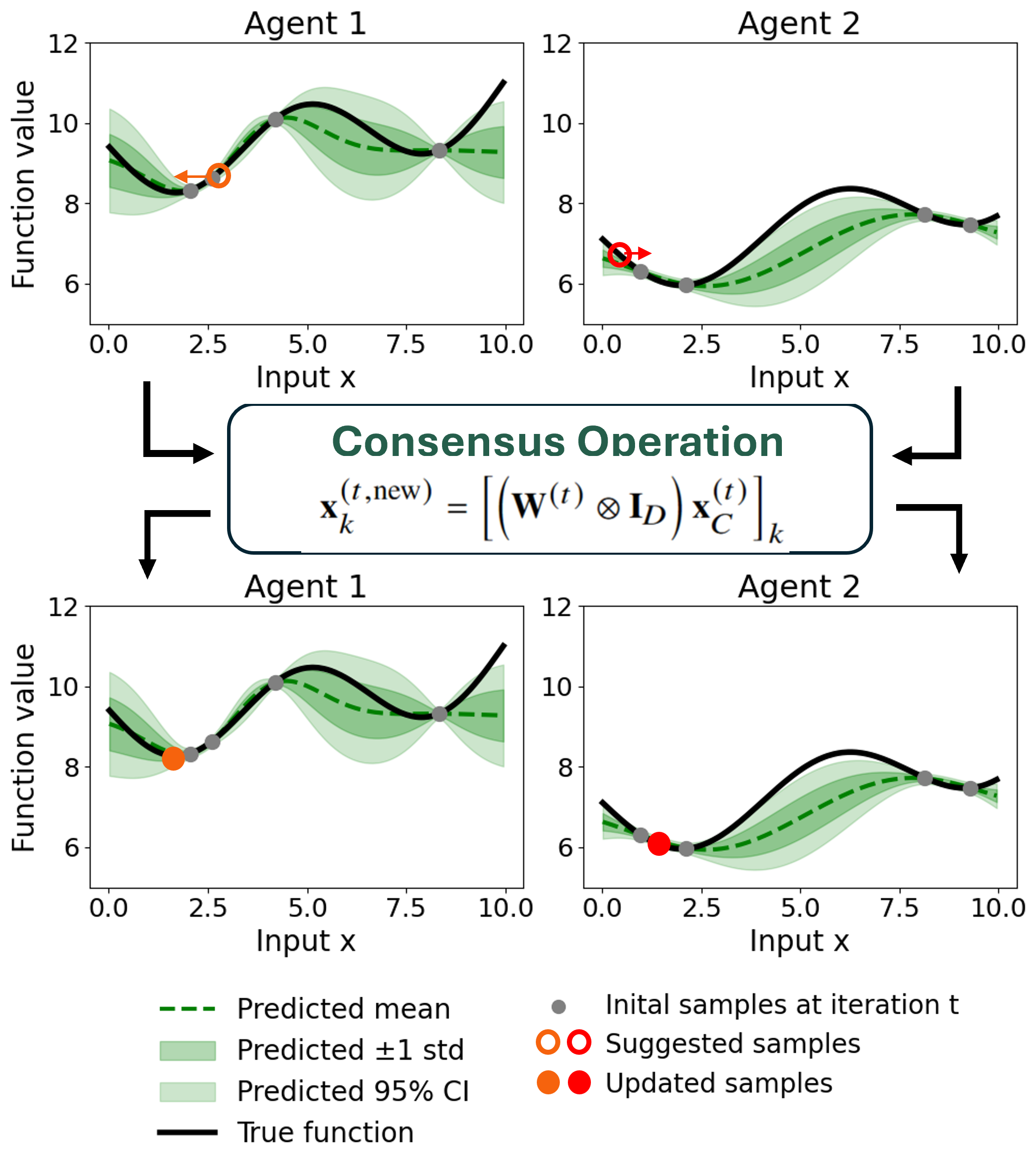}
    \caption{
    Consensus operation in collaborative Bayesian optimization. {\normalfont Agents first propose samples independently, then update their decisions via a weighted consensus based on the matrix \( \mathbf{W}^{(t)} \).}
    }
    \label{fig:col_operation}
\end{figure}

However, instead of immediately evaluating these proposals, the algorithm introduces a consensus step to enable collaboration. All agents' proposals \( \mathbf{x}_k^{(t)} \) are aggregated via a consensus mechanism defined by a doubly stochastic matrix \( \mathbf{W}^{(t)} \in \mathbb{R}^{K \times K} \). 
The updated, consensus-informed design for each agent is computed as the following: 

\begin{equation}
    \mathbf{x}_k^{(t,\text{new})}
    =
    \left[
    \left( \mathbf{W}^{(t)} \otimes \mathbf{I}_D \right)
    \mathbf{x}_C^{(t)}
    \right]_k,
\end{equation}
where \( \mathbf{x}_C^{(t)} \) concatenates the proposed designs from all agents, \( I_D \) is the \( D \times D \) identity matrix, \( \otimes \) denotes the Kronecker product, and \( [\cdot]_k \) extracts the \( k \)-th block corresponding to agent \( k \). The operation is illustrated in Figure~\ref{fig:col_operation}, showing a two-agent system where each agent proposes a sample independently and then updates it through consensus, blending local decisions with input from the other agent to coordinate on shared design variables.

Collaboration starts with full uniform averaging and gradually transitions towards independent optimization. Specifically, the consensus matrix is initialized as:

\begin{equation}
    \mathbf{W}^{(0)} = \frac{1}{K} \mathbf{1}_{K \times K},
\end{equation}
where \( \mathbf{1}_{K \times K} \) is a matrix of ones, representing equal weighting of all agents' proposals when initialized. As optimization progresses, the consensus matrix evolves according to:

\begin{equation}
    \mathbf{W}^{(t+1)} = \mathbf{W}^{(t)} + \frac{1}{TK} \left[ (K-1) \mathbf{I}_K - \mathbf{1}_{K \times K} \right],
\end{equation}
where \( T \) is the total number of iterations. This update increases the diagonal elements \( w_{kk} \) while decreasing the off-diagonal elements \( w_{kj} \) for \( k \neq j \), effectively reducing the influence of other agents over time. In the limit, \( \mathbf{W}^{(t)} \) converges to the identity matrix \( \mathbf{I}_K \), enabling agents to focus solely on their own objectives in later stages. The intuition behind this transitional design is that in early optimization stages, each agent's surrogate model is trained on limited data and benefits from the information pooled from other agents. As each agent accumulates more observations and constructs higher-quality surrogate models, the optimization naturally shifts towards agent-specific refinement.

\begin{figure}[H]
    \centering
    \includegraphics[width=1\linewidth]{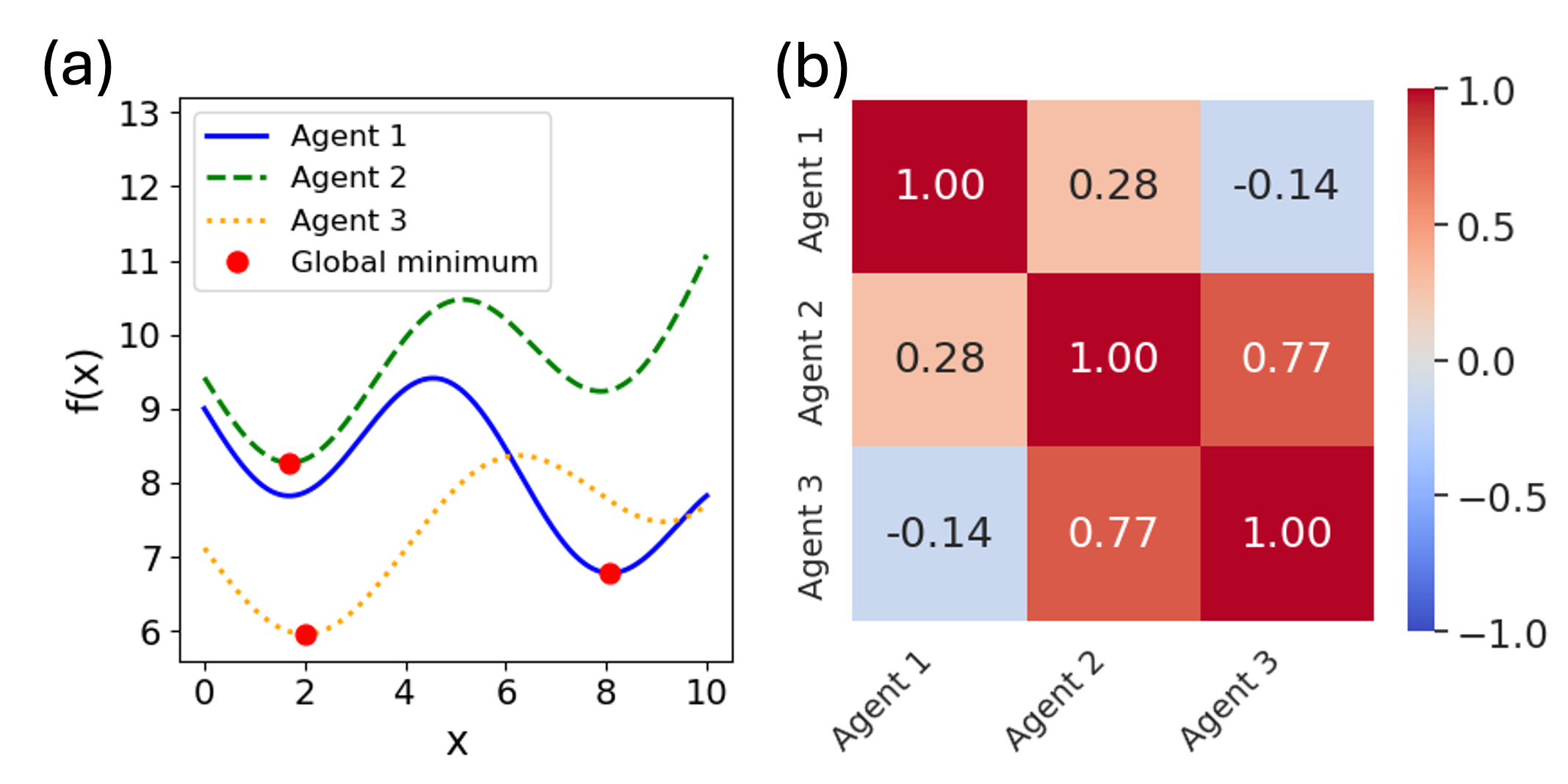}
    
    \caption{Illustration of function heterogeneity in a three-agent system. 
    {\normalfont (a) Agent-specific objective functions \(f_i(x)\), whose global minima lie in different regions of the domain. 
    (b) Pairwise similarity shown as a correlation matrix.}}
    \label{fig:heterogeneous}
\end{figure}

This consensus-based collaborative BO framework improves sample efficiency by coordinating exploration across agents while preserving task-specific goals. However, it relies on restrictive assumptions. It assumes agents optimize similar objectives with optima in the same region, overlooking functional heterogeneity. When objectives are misaligned or weakly correlated, uniform information sharing can cause negative transfer and slow convergence. As shown in Figure~\ref{fig:heterogeneous}, which depicts a three-agent system, Agents 2 and 3 are highly correlated (\(r=0.77\)), whereas Agents 1 and 3 are nearly uncorrelated (\(r=-0.14\)) with distinct optima. The framework further assumes equal evaluation budgets, ignoring resource asymmetries that lead to inefficient sampling and unbalanced contributions. It also neglects partially shared input spaces, where coordination is possible only over subsets of design variables. These limitations highlight the need for a more flexible, resource-aware, and heterogeneity-adaptive framework for decentralized multi-agent design.

\section{Adaptive Resource-aware COllaborative Bayesian Optimization (ARCO-BO)}
\label{sec:ARCO-BO}

We propose an Adaptive Resource-aware Collaborative Bayesian Optimization (ARCO-BO) framework to enhance collaborative BO in the presence of task heterogeneity, resource disparity, and partially shared input spaces. 
ARCO-BO extends the consensus-based collaborative BO by introducing three key mechanisms:
(1) Similarity and optima-aware dynamic consensus weighting, which promotes targeted knowledge exchange by adaptively weighting contributions from functionally related agents. 
(2) Budget-aware asynchronous sampling, which adjusts each agent’s sampling frequency based on its evaluation budget, ensuring efficient resource utilization.
(3) Partial input space sharing, which allows agents to collaborate even when their design spaces only partially overlap.
The overall workflow of ARCO-BO is summarized in Algorithm~\ref{alg:ARCO-BO}.

\begin{algorithm}[htbp]
\caption{ARCO-BO: Adaptive Resource-aware Collaborative Bayesian Optimization}
\label{alg:ARCO-BO}
\textbf{Given:} Initial data $\mathcal{D}_i = \{(\mathbf{x}_{ij}, y_{ij})\}_{j=1}^{n_i}$ for each agent $i \in \{1, \ldots, K\}$, objective functions $f_i$, budgets $B_i$, search domain $\mathcal{X}$ \\
\textbf{Goal:} Minimize each $f_i$ under budget constraints

\vspace{0.5em}
\textbf{Initialize:} Fit GP$_i$ using $\mathcal{D}_i$ for each agent.  
Let $B_{\max} = \max_i B_i$ and define update interval $\tau_i = \lfloor B_{\max} / B_i \rfloor$.  
Set iteration counter $t = 0$

\vspace{0.5em}
\textbf{While} any $B_i > 0$ and $t < T$ \textbf{do}:
\begin{enumerate}
    \item \textbf{Acquisition:}  
    For agent $i$, if $t \bmod \tau_i = 0$ and $B_i > 0$, select:
    \[
    \mathbf{x}_i^{(t)} = \arg\max_{\mathbf{x} \in \mathcal{X}} \alpha_i(\mathbf{x}; \text{GP}_i)
    \]
    
    \item \textbf{Consensus:}  
    Compute similarity matrix $\mathbf{S}^{(t)}$ and construct consensus weights $\mathbf{W}^{(t)}$  
    Update shared inputs:
    \[
    \mathbf{x}_{\text{shared}}^{(t)} \leftarrow \mathbf{W}^{(t)} \cdot 
    \begin{bmatrix}
    \mathbf{x}_{1,\text{shared}}^{(t)} \\
    \vdots \\
    \mathbf{x}_{K,\text{shared}}^{(t)}
    \end{bmatrix}
    \]

    \item \textbf{Evaluation and Update:}  
    For each agent $i$:
    \begin{itemize}
        \item Form input $\tilde{\mathbf{x}}_i^{(t)} = [\mathbf{x}_{\text{shared}}^{(t)}, \mathbf{x}_{i,\text{private}}^{(t)}]$
        \item Evaluate: $y_i^{(t)} = f_i(\tilde{\mathbf{x}}_i^{(t)})$
        \item Update data: $\mathcal{D}_i \leftarrow \mathcal{D}_i \cup \{(\tilde{\mathbf{x}}_i^{(t)}, y_i^{(t)})\}$
        \item Retrain GP$_i$, update budget: $B_i \leftarrow B_i - 1$
    \end{itemize}

    \item \textbf{Increment:} $t \leftarrow t + 1$
\end{enumerate}
\textbf{End while}

\vspace{0.5em}
\textbf{Output:} Final datasets $\mathcal{D}_i$ and solutions $\hat{\mathbf{x}}_i$ (best in $\mathcal{D}_i$) for each agent
\end{algorithm}

\subsection{Similarity- and Optima-Aware Dynamic Consensus Weighting}

Previous collaborative BO via consensus strategy \cite{yue2025collaborative} predominantly assumes uniform consensus weights, treating all agents as equally informative regardless of differences in their underlying objectives. While this assumption simplifies implementation, it is inherently limiting in heterogeneous settings where agents optimize related but non-identical tasks. Blindly aggregating information across dissimilar agents can degrade surrogate fidelity, introduce misleading guidance, and ultimately hinder optimization convergence. To address this limitation, ARCO-BO introduces a similarity- and optima-aware consensus mechanism that dynamically adjusts the influence of each agent based on both global functional similarity and local alignment of predicted optima. This approach ensures that information transfer is targeted and beneficial, promoting collaboration only where it enhances optimization performance.

Let $K$ denote the number of agents, each maintaining a local GP surrogate model $\text{GP}_k$. To quantify inter-agent similarity, ARCO-BO computes behavior-based embeddings derived from each model’s predictive mean over a shared input grid. Specifically, let $X_{\text{test}} \in \mathbb{R}^{N \times d}$ be a common test set sampled via Latin Hypercube Sampling within the design bounds.  Following established heuristics in surrogate modeling and design of computer experiments, we set $N = 50d$, which provides sufficient resolution for model comparison in moderate dimensions~\cite{wang2016bayesian,forrester2008engineering}. For each agent $k$, the predictive mean at each test input is computed to form a vector:
\begin{equation}
    \boldsymbol{\mu}_k = \mathbb{E}[\text{GP}_k(X_{\text{test}})],
\end{equation}
and its predicted minima location is determined by:
\begin{equation}
    \mathbf{x}^*_k = \arg \min_{x \in X_{\text{test}}} \boldsymbol{\mu}_k.
\end{equation}
If there happens to be more than one point with the same minimum value, we choose the one that appears first in the test set.

Stacking $\boldsymbol{\mu}_k$ across all agents yields the behavior matrix $\mathbf{M} \in \mathbb{R}^{K \times N}$, while stacking $\mathbf{x}^*_k$ yields the minima matrix $\mathbf{X}^* \in \mathbb{R}^{K \times d}$. These matrices encode both global predictive behavior and local optimal regions for all agents.

The optima-aware similarity between agents $i$ and $j$ is then defined as:

\begin{equation}
    s_{ij}
    =
    s_{ij}^{\text{Pearson}}
    \times
    s_{ij}^{\text{Proximity}},
\end{equation}
This multiplicative formulation ensures that high similarity is assigned only when both global trends and local optima are aligned. If either component is low, their product is suppressed, acting as a gating mechanism that avoids misleading collaboration. By enforcing joint agreement, the multiplicative form conservatively identifies agent pairs for which information sharing is both reliable and beneficial.

$s_{ij}^{\text{Pearson}}$ captures global similarity in functional behavior using the Pearson correlation coefficient, which measures the linear alignment between the predictive means of agents \( i \) and \( j \). Let \( \mu_{i,n} \) denote the \( n \)-th element of agent \( i \)’s predictive mean vector \( \boldsymbol{\mu}_i \), and let
\[
\bar{\mu}_i = \frac{1}{N} \sum_{n=1}^{N} \mu_{i,n}
\]
be the average predictive value over the test set. The Pearson correlation coefficient is then computed as:
\begin{equation}
\rho_{ij} =
\frac{ \sum_{n=1}^{N} (\mu_{i,n} - \bar{\mu}_i)(\mu_{j,n} - \bar{\mu}_j) }
{ \sqrt{ \sum_{n=1}^{N} (\mu_{i,n} - \bar{\mu}_i)^2 } \sqrt{ \sum_{n=1}^{N} (\mu_{j,n} - \bar{\mu}_j)^2 } },
\quad
s_{ij}^{\text{Pearson}} = \frac{ \rho_{ij} + 1 }{2 }.
\end{equation}
This normalization maps the Pearson correlation from its original range of \([-1, 1]\) to \([0, 1]\), ensuring non-negative similarity scores that are compatible with weighting in the consensus framework.

The second term, $s_{ij}^{\text{Proximity}}$, encodes minima proximity similarity, quantifying how close the predicted optimal designs are:

\begin{equation}
    s_{ij}^{\text{Proximity}}
    =
    \exp \left( - \lambda_p \| \mathbf{x}^*_i - \mathbf{x}^*_j \|^2 \right),
\end{equation}
where $\lambda_p$ is a tunable hyperparameter that controls the sensitivity of the similarity score to the distance between predicted minima. This exponential formulation ensures that agents with nearby optima are weighted more strongly in the consensus.

In our implementation, we set $\lambda_p$ such that $s_{ij}^{\text{Proximity}} = 0.1$ when the distance between predicted optima is 10\% of the input domain range $\Delta = \max(\mathbf{x}) - \min(\mathbf{x})$:

\begin{equation}
    \lambda_p = \frac{-\ln(0.1)}{(0.1 \Delta)^2}.
\end{equation}
The 10\% threshold is an empirically motivated choice based on initial experiments and reflects a moderate tolerance for optima misalignment. However, this threshold can be adjusted depending on the problem context and desired sensitivity to proximity. In general, $\lambda_p$ should be set to reflect the scale at which optima misalignment becomes detrimental to collaborative inference.

\begin{figure}[H]
    \centering
    \includegraphics[width=1\linewidth]{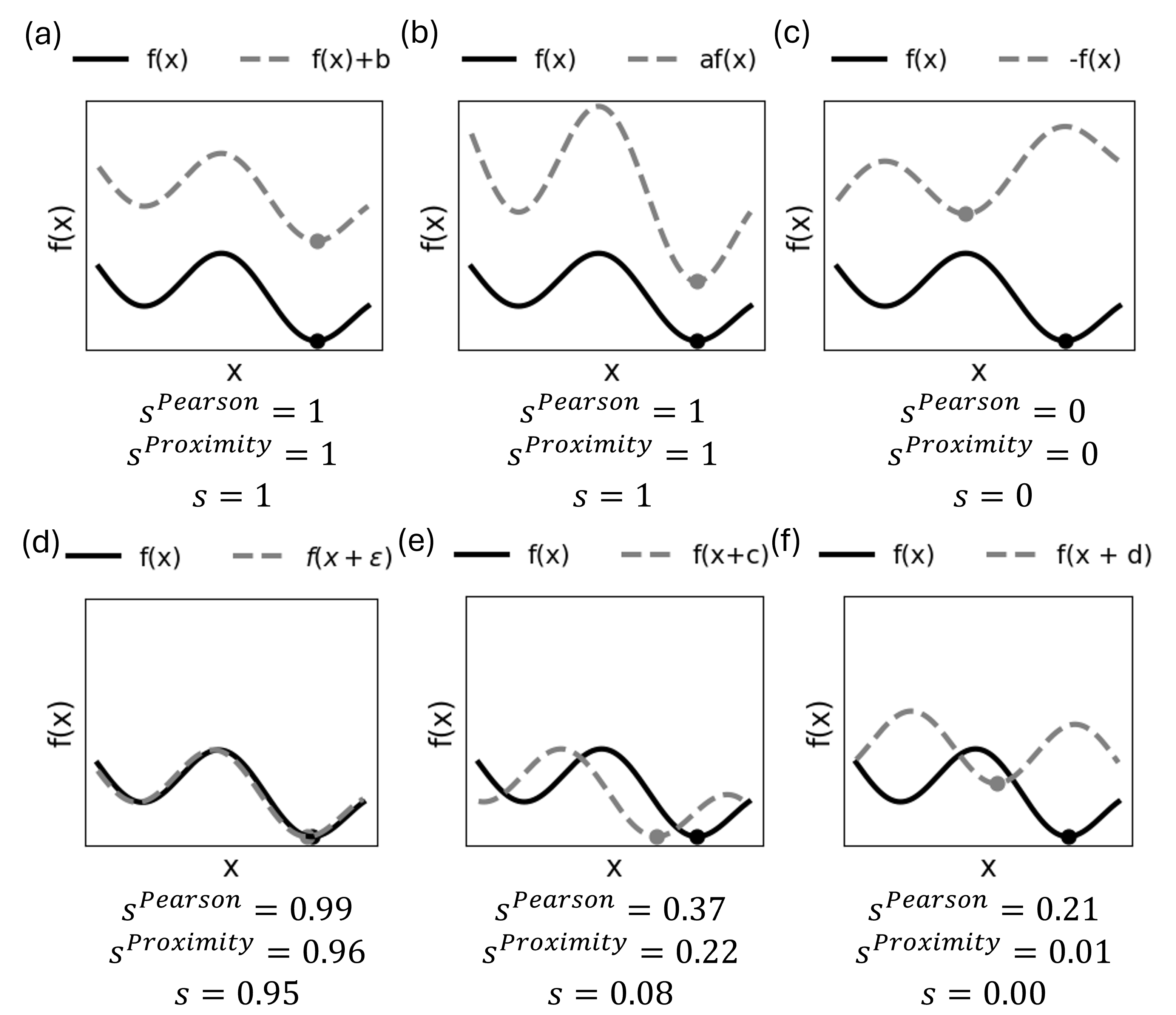}
    \caption{
    Illustration of function transformations and their impact on similarity.
    Each panel compares a base function $f(x)$ (solid black) with a transformed variant (dashed gray). {\normalfont
    (a) Vertical shift: $f(x) + b$ preserves both shape and minimizer location.
    (b) Amplitude scaling: $af(x)$ changes magnitude but keeps the same shape and minimizer.
    (c) Sign flip: $-f(x)$ inverts the shape and shifts the minimizer.
    (d) Small input shift: $f(x + \varepsilon)$ retains high similarity in both shape and minimizer location.
    (e) Moderate input shift: $f(x + c)$ preserves shape but misaligns the minimizer.
    (f) Large input shift: $f(x + d)$ leads to low shape and proximity similarity.
    Pearson correlation measures shape similarity, while proximity reflects alignment between minimizers.
    }}
    \label{fig:illu_sim}
\end{figure}

Together, these yield a symmetric similarity matrix $\mathbf{S}^{(t)} \in \mathbb{R}^{K \times K}$ that captures both global predictive consistency and alignment in locally optimal regions. To illustrate the motivation behind our metric, Figure~\ref{fig:illu_sim} shows how different function transformations impact shape similarity and optima proximity. Each subplot compares a base function \( f(x) \) (solid black) with a transformed variant (dashed gray). Transformations like vertical shift and amplitude scaling (Figure~\ref{fig:illu_sim}a–b) preserve both shape and minimizer location, resulting in high similarity. In contrast, sign flips (Figure~\ref{fig:illu_sim}c) or input shifts (Figure~\ref{fig:illu_sim}d–f) degrade similarity by altering the function shape, the optima location, or both. These examples highlight the need for a similarity metric that considers both global trends and local optima. Our multiplicative formulation conservatively gates information sharing, emphasizing collaboration only when both aspects are well aligned.

To enable dynamic, similarity-aware collaboration, ARCO-BO constructs a time-varying consensus matrix:
\begin{equation}
    \mathbf{W}^{(t)} = \gamma(t) \cdot \mathbf{S}^{(t)} + \left(1 - \gamma(t)\right) \cdot \mathbf{I},
\end{equation}
where \( \mathbf{S}^{(t)} \in \mathbb{R}^{K \times K} \) is a non-negative similarity matrix, and \( \mathbf{I} \) is the identity matrix. The scalar weighting function
\begin{equation}
    \gamma(t) = \exp\left(-\frac{\alpha t}{T}\right)
\end{equation}
controls the transition from early-stage collaboration (\( \gamma(0) = 1 \)) to late-stage independent optimization (\( \gamma(T) \approx 0 \)). Here, \( T \) denotes the total number of BO iterations, and \( \alpha \) is a tunable hyperparameter that governs the decay rate. Larger values of \( \alpha \) lead to faster reduction in collaboration, while smaller values preserve stronger consensus for a longer duration.

To ensure that \( \mathbf{W}^{(t)} \) defines a valid consensus mechanism, it is normalized using Sinkhorn normalization~\cite{sinkhorn1964relationship}, which iteratively rescales the rows and columns of a non-negative matrix to make it doubly stochastic. The normalization is applied as:
\begin{equation}
    \mathbf{W}^{(t)} \leftarrow \text{diag}(\mathbf{r})^{-1} \mathbf{W}^{(t)} \, \text{diag}(\mathbf{c})^{-1},
\end{equation}
where \( \mathbf{r} \) and \( \mathbf{c} \) are row and column scaling vectors, updated alternately until convergence.

This similarity- and optima-aware consensus strategy enables ARCO-BO to integrate information across agents only when it is mutually beneficial, thereby enhancing sample efficiency and improving convergence robustness in heterogeneous multi-agent optimization settings.

\subsection{Budget-aware Asynchronous Sampling}

To realize the agent-specific sampling intervals \( \tau_i \) introduced in of Algorithm~\ref{alg:ARCO-BO}, we incorporate a budget-aware asynchronous sampling mechanism.

In real-world collaborative optimization, agents often operate under heterogeneous resource constraints. For instance, one agent might perform high-fidelity physical experiments that are expensive and time-consuming, while another relies on low-fidelity simulations that are faster and cheaper to evaluate. Similarly, some agents may be limited by equipment availability or operational cost, while others can afford frequent queries due to lower evaluation overhead. Despite this variability, most existing collaborative BO methods assume a uniform sampling rate, where all agents carry out the same number of iterations or participate in collaborative updates at the same frequency. This assumption can lead to inefficient use of resources, as fast agents are forced to wait for slower ones, and costly evaluations are scheduled just as frequently as cheap ones. ARCO-BO addresses this by introducing a budget-aware asynchronous sampling mechanism. Specifically, each agent $i$ is assigned a sampling interval $\tau_i$ based on its relative budget:

\begin{equation}
    \tau_i = \left\lfloor \frac{B_{\max}}{B_i} \right\rfloor,
\end{equation}
where $B_i$ is the total budget for agent $i$ and $B_{\max}$ is the maximum budget among all agents. This means agents with larger budgets sample more frequently, while those with smaller budgets sample less often, helping them conserve their limited evaluations for the most important points.

Agent~$i$ proposes a new point every $\tau_i$ iterations when it still has remaining budget. Otherwise, it skips that iteration. This asynchronous sampling strategy allows each agent to operate at a frequency proportional to its allocated budget, enabling more efficient resource utilization and improved overall performance in collaborative optimization.

Currently, ARCO-BO assumes that the evaluation budgets $B_i$ are fixed and known in advance. However, the framework can be extended to accommodate dynamic budget adjustment by periodically updating $\tau_i$ based on remaining budgets or runtime constraints. Similarly, the model could be adapted to account for varying evaluation costs by defining an effective budget in terms of cumulative cost rather than number of evaluations. We leave these extensions for future work.

\subsection{Partial Input Space Sharing}

Building on our dynamic consensus mechanism of Algorithm~\ref{alg:ARCO-BO}, we now handle cases where agents share only a subset of the design variables.

In many collaborative optimization problems, agents may have access to only a subset of the global design variables due to factors such as unmeasurable parameters in experiments or privacy constraints. As a result, their input spaces only partially overlap. Conventional BO frameworks often assume a fully shared input space, which limits their applicability in such heterogeneous settings. ARCO-BO addresses this by enabling partial input space sharing within its consensus updates. Each agent’s input is partitioned into shared and private components:

\begin{equation}
    \mathbf{x}_i = \begin{bmatrix}
    \mathbf{x}_{\text{shared},i} \\
    \mathbf{x}_{\text{private},i}
    \end{bmatrix},
\end{equation}
where $\mathbf{x}_{\text{shared},i} \in \mathbb{R}^{d_s}$ are variables shared across agents, and $\mathbf{x}_{\text{private},i} \in \mathbb{R}^{d_p}$ are agent-specific, such that $d_s + d_p = d$. Note that $d_s$ and $d_p$ may vary by agent.

During consensus updates, only the shared subspace is aggregated:

\begin{equation}
    \mathbf{x}_{\text{shared}}^{(t+1)} = \mathbf{W}^{(t)} \mathbf{x}_{\text{shared}}^{(t)},
\end{equation}
while private variables remain unchanged:

\begin{equation}
    \mathbf{x}_{\text{private},i}^{(t+1)} = \mathbf{x}_{\text{private},i}^{(t)}, \quad \forall i.
\end{equation}

The final design combines both components:

\begin{equation}
    \mathbf{x}_i^{(t+1)} = \begin{bmatrix}
    \mathbf{x}_{\text{shared},i}^{(t+1)} \\
    \mathbf{x}_{\text{private},i}^{(t+1)}
    \end{bmatrix}.
\end{equation}

This partial-sharing scheme ensures agents exchange only the overlapping dimensions, preserving privacy of their private inputs while still leveraging joint exploration.

\section{Evaluation Metrics}
\label{sec:exp}

In this section, we introduce the quantitative metrics used to evaluate the performance of multi-agent Bayesian optimization methods. In ARCO-BO, each of the $K$ agents performs one function evaluation per iteration, so one global iteration corresponds to up to $K$ parallel function calls. This execution model reflects a parallel deployment setting, where agents operate independently and evaluate simultaneously without coordination.

Given that the objective is to identify the global optimum for each individual agent, the evaluation metrics are designed to assess performance across all agents collectively.

\begin{enumerate}
    \item \textbf{Normalized Final Regret.}
    To account for differences in objective function scales across agents, we evaluate performance using the normalized final regret \cite{grosnit2021we}:
    \begin{equation}
        \bar{r}^{\text{final}} = \frac{1}{K} \sum_{i=1}^{K}
        \frac{f_i(\mathbf{x}_i^*) - \min_{t = 1,\dots,T} f_i(\mathbf{x}_i^{(t)})}{f_i^{\max} - f_i^{\min}},
    \end{equation}
    where \( f_i(\mathbf{x}_i^*) \) is the known global minimum for agent \( i \), and \( f_i^{\max} \), \( f_i^{\min} \) denote the maximum and minimum of the true function. A smaller value of \( \bar{r}^{\text{final}} \) indicates closer proximity to the optimum.

    \item \textbf{Normalized Area Under the Curve (AUC).}
    To evaluate early-stage convergence behavior, we compute the normalized area under the averaged convergence curve \cite{xie2025merge}, which tracks the best objective value found so far over time. Specifically,
    \begin{equation}
        \overline{\text{AUC}} = \frac{1}{K} \sum_{i=1}^{K} \frac{1}{N} \sum_{t=1}^{N}
        \frac{\bar{f}_i^{(t)} - f_i(\mathbf{x}_i^*)}{f_i^{\max} - f_i^{\min}},
    \end{equation}
    where \( \bar{f}_i^{(t)} \) is the mean of the best objective value found by agent \( i \) up to iteration \( t \), averaged across replicates, and \( N \) is the number of early-stage iterations. A smaller \( \overline{\text{AUC}} \) indicates faster convergence to the optimum during the initial phase of optimization. In our test cases, we use \( N =  0.1 T \) to represent the early-stage phase of optimization.

\end{enumerate}

All experiments were performed on a workstation running Ubuntu 20.04, equipped with an AMD EPYC 7413 processor (24 cores) and 64 GB of RAM. The code was implemented in Python 3.9 and executed on the CPU.

\section{Case Studies}
\label{sec:case_study}

In this section, we first present a 1D illustrative example demonstrating how our method works differently in contrast to conventional collaborative BO via consensus when optimizing heterogeneous functions. 
Next, we demonstrate the effectiveness of the proposed ARCO-BO method on a 2D example exhibiting three types of heterogeneity: task heterogeneity, resource disparity, and partially shared input spaces. This example highlights the advantages of ARCO-BO over applying separate BO independently to each agent.
Finally, we evaluate ARCO-BO on two high dimensional engineering benchmarks: the borehole function (8D) and the wing weight function (10D). These engineering problems are widely used in surrogate-assisted optimization and test ARCO-BO's ability to scale in terms of both consensus overhead and distributed sampling efficiency in higher dimensions.

It is important to note that although these benchmark functions are originally used as test cases for multi-fidelity designs \cite{chen2024latent}, we treat each fidelity level as a separate agent task in this study, each with its own objective and ground truth, rather than treating the highest fidelity function as a ground-truth, and the objective is to identify the optimal solution for each agent simultaneously.

\subsection{1D Illustrative Example}

To evaluate the robustness and collaborative dynamics of ARCO-BO, we consider a multi-agent optimization problem based on variants of the Sasena function \cite{lin2022probability}, where three agents each optimizes distinct but related functions with different global minimum locations. These variations emulate realistic heterogeneity arising from differences in modeling fidelity, operating conditions, or task formulations, which is common in engineering design and simulation. Details of the agent-specific functions, lower and upper bounds, budgets, and true optimal are provided in Table~\ref{tab:sasena}.
Each agent begins with three randomly sampled initial evaluations, which are kept consistent across all compared methods to ensure a fair comparison.

\begin{figure*}[h!]
    \centering
    \includegraphics[width=0.75\linewidth]{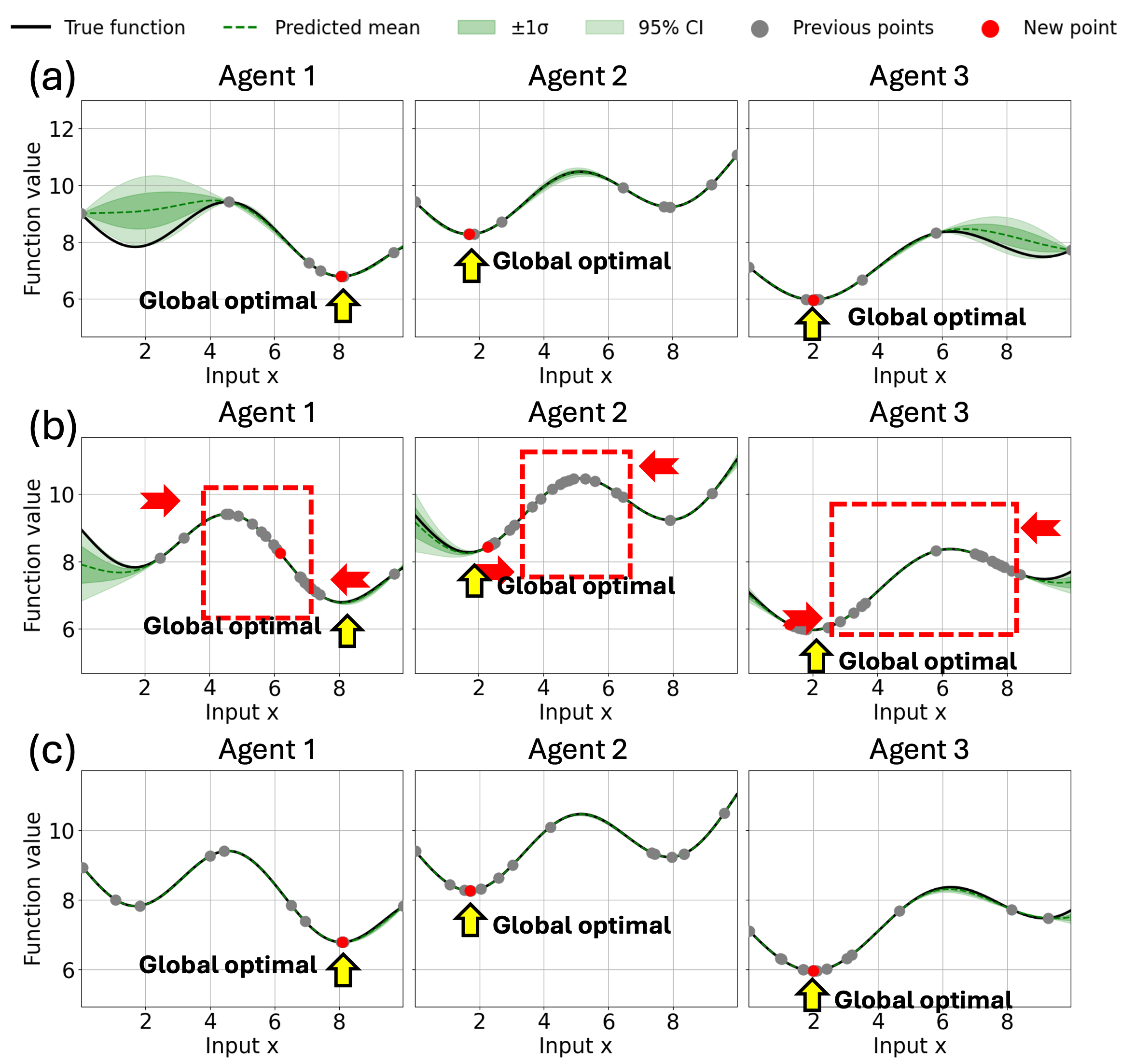}
    \caption{Comparison of surrogate model predictions across three optimization strategies at iteration 20 for three agents optimizing distinct variants of the Sasena function. {\normalfont (a) separate BO, (b) benchmark collaborative BO and (c) our proposed ARCO-BO.}}
    \label{fig:sasena-col}
\end{figure*}

\begin{figure*}[h!]
    \centering
    \includegraphics[width=0.9\linewidth]{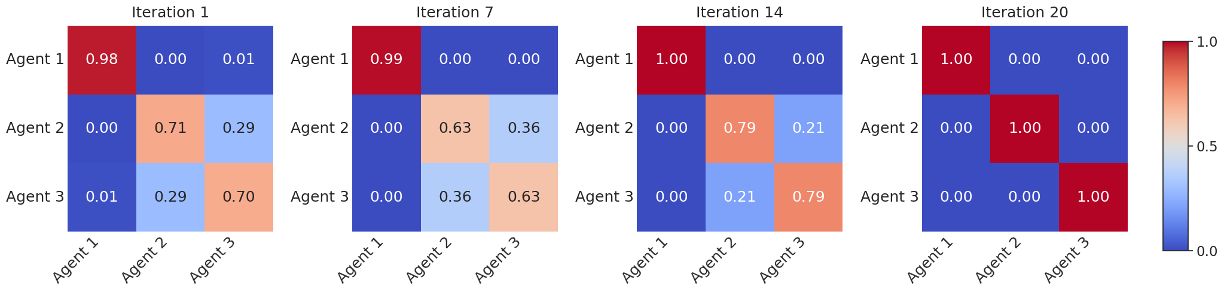}
    \caption{Progression of the learned inter-agent weights W under ARCO-BO at iterations 1, 7, 14, and 20.}
    \label{fig:W}
\end{figure*}

In this example, all Gaussian process surrogate models use a squared exponential (RBF) kernel, with fixed lengthscale $0.5$, signal variance $1.0$, and Gaussian observation noise with variance $10^{-6}$.
Figure~\ref{fig:sasena-col} illustrates the GP updates at the last iteration under three optimization strategies. Figure~\ref{fig:sasena-col}a shows a separate BO, where each agent optimizes independently without collaboration. While the predicted means generally align with the true functions, the models exhibit high uncertainty in unexplored regions. 
Figure~\ref{fig:sasena-col}b illustrates the behavior of an existing collaborative BO method based on uniform consensus, where agents share information equally regardless of task alignment. This uniform sharing often leads to incorrect collaboration, as conflicting information from agents with different objectives is combined without regard for compatibility. Consequently, agents are misled in their sampling decisions, resulting in suboptimal exploration and poor convergence. This issue is visually highlighted by the red boxes and arrows, which indicate significant deviations from the true optima.
In contrast, Figure~\ref{fig:sasena-col}c shows the results of ARCO-BO, which adaptively reweights inter-agent influence based on task similarity and differences in optimal solution locations. The four heatmaps (Figure~\ref{fig:W}) depict the learned inter-agent weight matrix \(W\), where each cell \((i,j)\) is the weight agent \(i\) assigns to agent \(j\). Because Agent~1’s optimum is far from those of Agents~2 and~3, their pairwise similarity is near zero; consequently ARCO-BO suppresses the cross-weights \(W_{12}\) and \(W_{13}\) (and symmetrically \(W_{21}\), \(W_{31}\)) toward zero. By contrast, Agents~2 and~3 are similar, so ARCO-BO initially assignscross-weights \(W_{23}\) and \(W_{32}\) to facilitate sharing; as both agents approach their optima, these weights taper and \(W\) sharpens toward a near-diagonal structure, evidencing reduced cross-agent influence and increasing self-reliance. This targeted collaboration enables agents to focus sampling near their true optima while avoiding misleading guidance from incompatible tasks.

\begin{figure*}[h!]
    \centering
    \includegraphics[width=0.95\linewidth]{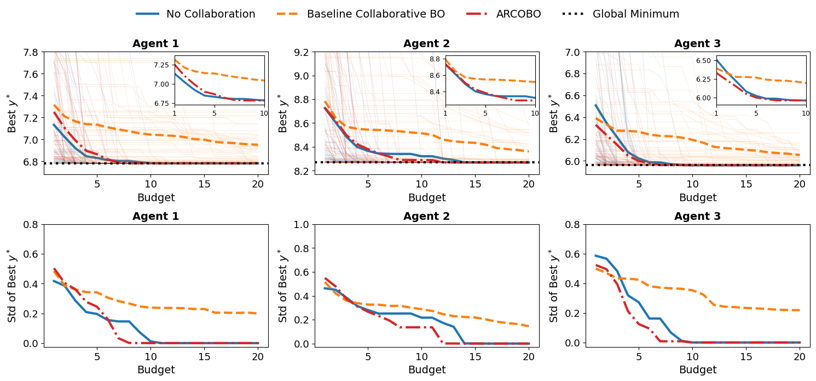}
    \caption{Convergence performance of ARCO-BO compared to baseline methods on the multi-agent Sasena problem, evaluated over 50 independent replicates.}
    \label{fig:sasena-results}
\end{figure*}

Figure~\ref{fig:sasena-results} further quantifies the benefits of ARCO-BO based on 50 independent replicates, each generated using different initial samples drawn for the agents. In the top row, each plot shows the mean best function value ($y^*$) over iterations, comparing ARCO-BO with conventional collaborative BO via consensus and separate BO. The results demonstrate that ARCO-BO enables faster convergence to each agent’s optimum. In contrast, the baseline collaborative BO fails to outperform separate BO and often does not reach the true optima, due to overly strong information sharing that misguides agents with divergent objectives. The bottom row shows the standard deviation of the best function values, indicating that ARCO-BO also achieves greater robustness in optimization performance. To further assess performance, we compute the normalized AUC and the normalized final regret, as summarized in Table~\ref{tab:Sasena-results}. Both separate BO and ARCO-BO are able to reach the global optima for all agents, while the baseline collaborative BO fails to do so. Among all methods, ARCO-BO achieves the fastest early convergence, highlighting the effectiveness of its adaptive and similarity-aware collaboration strategy.

\begin{table}[htbp]
\centering
\caption{Normalized AUC and normalized final regrets for the 1D Sasena problem, reported as mean ± standard deviation across 50 replicates.}
\label{tab:Sasena-results}
\begin{tabular}{lcc}
\toprule
\textbf{Method} & \textbf{AUC} & \textbf{Final Regret} \\
\midrule
Separate BO     & $0.1623 \pm 0.2048$ & $0.0000 \pm 0.0000$ \\
Benchmark CBO          & $0.1840 \pm 0.1872$ & $0.0483 \pm 0.0790$ \\
ARCO-BO                              & $0.1562 \pm 0.2017$ & $0.0000 \pm 0.0000$ \\
\bottomrule
\end{tabular}
\end{table}

\begin{table}[htbp]
\centering
\caption{Normalized AUC and normalized final regrets for the 2D Ackley problem in three different scenarios, reported as mean ± standard deviation across 50 replicates.}
\label{tab:ackley-results-summary}
\begin{tabular}{clcc}
\toprule
\textbf{Scen.} & \textbf{Method} & \textbf{AUC} & \textbf{Final Regret} \\
\midrule
1 & Separate BO        & $0.2784 \pm 0.1483$ & $0.0169 \pm 0.0159$ \\
  & Benchmark CBO      & $0.2050 \pm 0.0870$ & $0.0929 \pm 0.0669$ \\
  & ARCO-BO             & $0.2008 \pm 0.0932$ & $0.0145 \pm 0.0158$ \\
\midrule
2 & Separate BO        & $0.2770 \pm 0.1486$ & $0.0143 \pm 0.0159$ \\
  & ARCO-BO             & $0.1992 \pm 0.0933$ & $0.0125 \pm 0.0158$ \\
\midrule
3 & Separate BO        & $0.2784 \pm 0.1483$ & $0.0186 \pm 0.0186$ \\
  & ARCO-BO             & $0.1861 \pm 0.0975$ & $0.0145 \pm 0.0158$ \\
\bottomrule
\end{tabular}
\end{table}

\begin{figure*}[h!]
    \centering
    \includegraphics[width=0.95\linewidth]{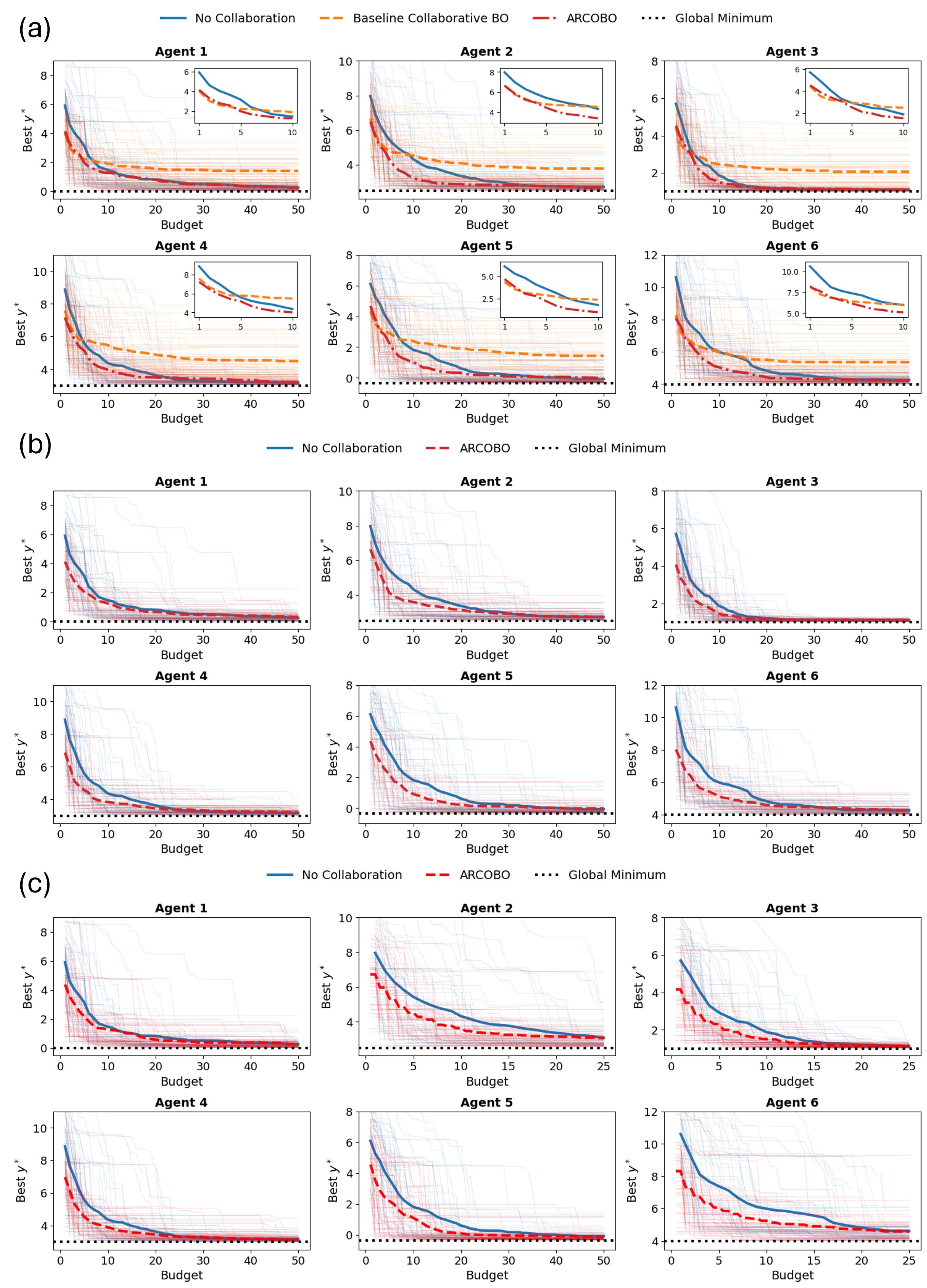}
    \caption{Convergence performance across 50 replicates for ARCO-BO and separate BO on the multi-agent Ackley problem with (a) same budgets and fully shareable input, (b) different budgets for each agents, (c) different shareable inputs.}
    \label{fig:ackley-budget-share}
\end{figure*}

\subsection{2D Illustrative Example}

We consider a multi-agent problem based on variations of the 2D Ackley function \cite{ackley2012connectionist}, a widely used benchmark known for its numerous local minima and challenging non-convex landscape.
We design three scenarios to assess ARCO-BO’s capabilities: (1) agents have the same evaluation budgets with both input variables fully shareable; (2) agents have different evaluation budgets with both input variables shareable, simulating heterogeneous resource availability; and (3) agents have the same budgets but only one input variable is shareable, reflecting partial input sharing constraints. In all cases, ARCO-BO is compared against separate BO and, where applicable, a benchmark collaborative BO via consensus.

Detailed information about the six agents is listed in Table~\ref{tab:ackley}. In the second scenario, agents 2, 3, and 6 are assigned only half of the original evaluation budgets to reflect resource differences. Each scenario is evaluated across 50 independent replicates, and we report both the normalized AUC and the normalized final regret for all agents (Table~\ref{tab:ackley-results-summary}). We use an RBF kernel with fixed lengthscale $0.5$, signal variance $1.0$, and observation noise variance $10^{-6}$.

In scenario 1 (Figure~\ref{fig:ackley-budget-share}a), where agents have equal evaluation budgets and both input variables are fully shareable, ARCO-BO achieves rapid convergence. Compared to the benchmark collaborative BO, which often fails to converge to each agent’s local optimum due to detrimental information transfer, ARCO-BO demonstrates clear advantages by enabling agents to collaborate selectively only when it is mutually beneficial.

In scenario 2 (Figure~\ref{fig:ackley-budget-share}b), where agents have different evaluation budgets but both input variables remain shareable, ARCO-BO maintains robust optimization performance despite the heterogeneous resource distribution. Notably, separate BO struggles under this setting because agents with low budgets cannot adequately explore their objective landscapes, which leads to the convergence on suboptimal solutions (agent 2, 3 and 6 in Figure~\ref{fig:ackley-budget-share}b). In contrast, ARCO-BO utilizes knowledge transferred from higher-budget agents to guide low-budget agents towards better solutions, leading to improvements in convergence speed and final regret compared to baselines.

In scenario 3 (Figure~\ref{fig:ackley-budget-share}c), where agents have the same budgets but only one input variable is shareable, ARCO-BO continues to demonstrate its advantages over separate BO. While partial input sharing limits the extent of collaboration, ARCO-BO’s adaptive consensus mechanism ensures that agents only share useful and consistent information along the shared dimensions. As a result, it still outperforms independent BO, which lacks collaborative benefits.

Overall, these results demonstrate that ARCO-BO’s adaptive collaboration strategy works with varying input sharing constraints and resource heterogeneity. It provides an efficient optimization framework for scenarios where traditional BO frameworks struggle due to limited budgets or incomplete information sharing.

\subsection{Example on High-Dimensional Engineering Benchmarks}

\begin{figure*}[h!]
    \centering
    \includegraphics[width=\linewidth]{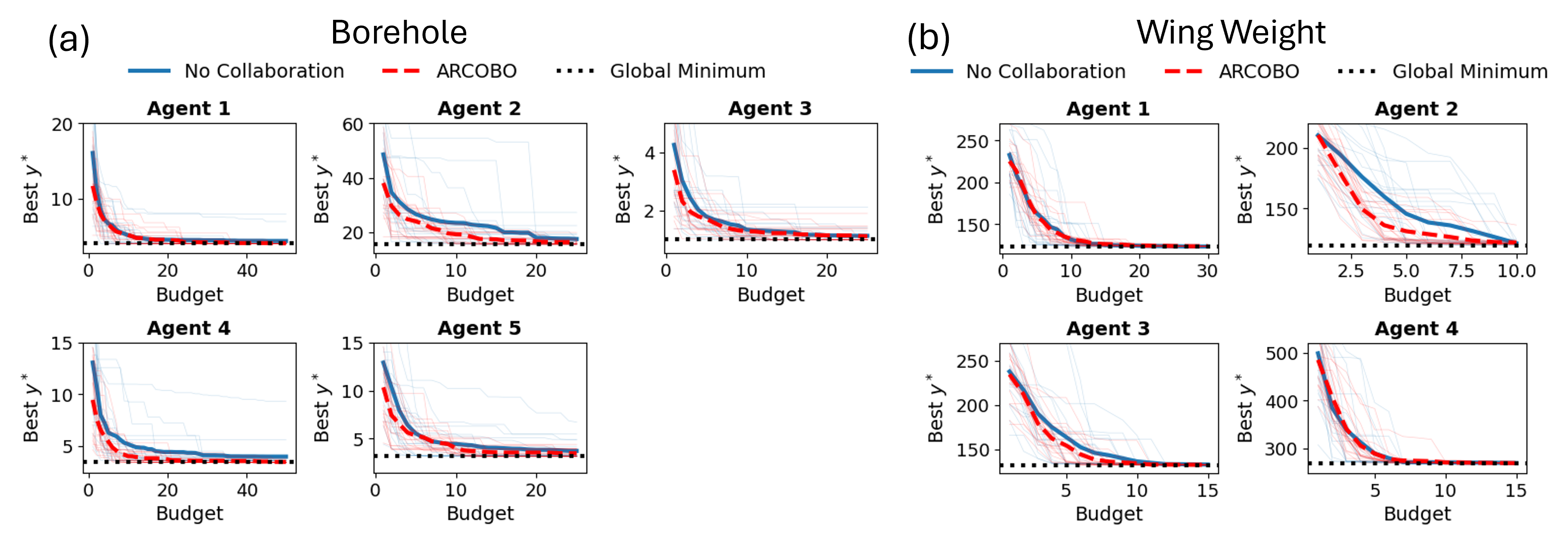}
    \caption{Convergence performance across 20 replicates for ARCO-BO and separate BO on the (a) Borehole problem, and (b) Wing Weight problem.}
    \label{fig:high_dimension}
\end{figure*}

We performed 20 independent replicates for both the Borehole and Wing Weight problems under a multi-agent setting. These two functions are both modified from \cite{chen2024latent}. The results are summarized in Table~\ref{tab:high-dimension} and visualized in Fig.~\ref{fig:high_dimension}. In each benchmark, agents operate under different evaluation budgets, and only a subset of the input space is shared across agents. Full specifications of the agent-specific functions, budget allocations, and shared input dimensions are provided in Table~\ref{tab:borehole} and Table~\ref{tab:wingweight}.

Across both benchmark problems, ARCO-BO consistently outperforms separate BO in terms of both early convergence and final solution quality, as shown in Table~\ref{tab:high-dimension}. Figure~\ref{fig:high_dimension}a illustrates these gains on the Borehole benchmark. For example, Agents~2, 3, and~5 operate under reduced evaluation budgets, which would typically limit their ability to reach the global optimum. However, with ARCO-BO, these agents are able to benefit from collaborative information sharing with higher-budget peers, allowing them to converge effectively despite fewer evaluations. A similar trend is observed in Figure~\ref{fig:high_dimension}b for the Wing Weight benchmark, particularly for Agent~2 where ARCO-BO shows significant improvement over separate BO. Without collaboration, the agent lacks sufficient information to escape poor regions of the design space. ARCO-BO, by contrast, enables faster convergence by leveraging inter-agent knowledge to guide optimization across heterogeneous settings.

The superior performance of ARCO-BO across these engineering benchmarks stems from its ability to adaptively coordinate learning across heterogeneous agents. Using similarity-aware consensus, it avoids negative transfer from dissimilar tasks, which improves convergence. Its budget-aware asynchronous sampling helps agents with limited evaluations allocate resources more effectively. Partial input space sharing enables collaboration even when agents operate on different subsets of the design space. Together, these mechanisms let ARCO-BO exploit complementary information, increase sample efficiency, and deliver consistently better solutions than independent BO in complex multi-agent settings.

\begin{table}[htbp]
\centering
\caption{Normalized AUC and final regrets for Borehole and Wing Weight problems (mean ± std over 20 replicates).}
\label{tab:high-dimension}
\begin{tabular}{lccc}
\toprule
\textbf{Benchmark} & \textbf{Method} & \textbf{AUC} & \textbf{Final Regret} \\
\midrule
\multirow{2}{*}{Borehole}
& Separate BO & $0.0251 \pm 0.0178$ & $0.0018 \pm 0.0038$ \\
& ARCO-BO      & $0.0174 \pm 0.0101$ & $0.0008 \pm 0.0016$ \\
\midrule
\multirow{2}{*}{Wing Weight}
& Separate BO & $0.0807 \pm 0.0757$ & $0.0274 \pm 0.0576$ \\
& ARCO-BO      & $0.0471 \pm 0.0188$ & $0.0026 \pm 0.0050$ \\
\bottomrule
\end{tabular}

\end{table}

\section{Conclusion}
\label{sec:conclusion}

This work introduces ARCO-BO, a framework for multi-agent BO in heterogeneous settings. ARCO-BO addresses key challenges such as task heterogeneity, resource disparity, and partially shared input spaces, which are common in real-world collaborative scenarios such as distributed manufacturing, material discovery, and multi-fidelity design. The proposed framework integrates three core innovations to enable effective collaboration:

\begin{itemize}
    \item Similarity and optima-aware consensus mechanisms that adaptively regulate inter-agent influence based on surrogate model alignment.  
    \item A budget-aware sampling strategy that allocates evaluations according to each agent’s available resources.  
    \item A strategy to handle partially shared inputs that enables collaboration while preserving private design variables.  
\end{itemize}

Experiments on synthetic problems and engineering benchmarks show that ARCO-BO consistently outperforms independent BO and conventional collaborative BO via consensus. It converges faster and achieves lower regret under heterogeneous budgets and partially shared inputs, with low-budget agents gaining from targeted knowledge transfer by high-budget peers without being misled by irrelevant information.

These gains stem from ARCO-BO’s distinctive approach to information sharing, which sets it apart from existing multi-agent design strategies. Consensus-based collaborative BO methods coordinate decisions while avoiding raw data exchange, but their reliance on uniform averaging often ignores task heterogeneity and budget asymmetry. At the other extreme, multi-source data fusion methods combine all agents’ input–output data into a unified surrogate, but require full data sharing and degrade in performance under strong heterogeneity or privacy constraints.

Despite its advantages, ARCO-BO still assumes fixed evaluation budgets and depends on user-defined hyperparameters, such as the consensus decay rate, which may limit its adaptability in dynamic environments. Future extensions could incorporate adaptive budgeting strategies, self-tuning consensus mechanisms, and dynamic agent participation, as well as support for richer forms of input heterogeneity, including mixed discrete–continuous domains and agent-specific constraints. Beyond addressing these limitations, ARCO-BO also opens broader opportunities. Its emphasis on selective and resource-aware collaboration aligns naturally with settings where data sharing is constrained and evaluations are expensive, such as federated optimization, cooperative scientific design, and distributed engineering. Reinforcement learning offers another promising direction, where consensus updates could themselves be adapted through learned policies that regulate inter-agent influence over time, guide sampling with a long-term view of performance.


\appendix
\renewcommand{\thefigure}{A\arabic{figure}}
\renewcommand{\thetable}{A\arabic{table}}
\setcounter{figure}{0}
\setcounter{table}{0}

\section{Details of Test Functions}
\label{appendix:test_funcs}

This section presents detailed formulations and configurations for all benchmark functions used in our study. These include two illustrative examples—a 1D Sasena function (Table~\ref{tab:sasena}) and a 2D multi-fidelity Ackley function  (Table~\ref{tab:ackley}), as well as two engineering-inspired high-dimensional problems: the Borehole (Table~\ref{tab:borehole}) and Wing Weight (Table~\ref{tab:wingweight}) functions. For each case, we specify the agent-specific modifications, input bounds, sample budgets, and known optima. These functions collectively span a range of heterogeneities in input dimension, fidelity structure, and inter-agent similarity, serving to rigorously evaluate the proposed optimization framework.

\begin{table*}[h!]
\centering
\caption{1D illustrative example: Sasena function. Formulation, upper/lower bounds, initial sample size, budget, and true optimal value of each agent.}
\label{tab:sasena}
\begin{tabular}{p{1.5cm} p{5cm} p{2cm} p{2cm} p{1.5cm} p{2cm}}
\hline
Agent & Formulation & Upper/lower bound & Init. Sample & Budget & True optimal \\
\hline
1 &
$y_{1}(x) = -\sin x - \exp\left(\frac{x}{10}\right) + 10$ &
$0 \le x \le 10$ &
3 &
20 &
6.782 \\
2 &
$y_{2}(x) = -\sin(0.95x) - \exp\left(\frac{x}{50}\right) + 0.03(x-2)^2 + 10.3$ &
& 3 &
20 &
8.269 \\
3 &
$y_{3}(x) = -\sin(0.8x) - \exp\left(\frac{x}{50}\right) + 0.03(x-2)^2 + 8$ &
& 3 &
20 &
5.959 \\
\hline
\end{tabular}
\end{table*}

\begin{sidewaystable*}[h!]
\centering
\caption{2D illustrative example: Multi-fidelity Ackley function. Formulation, bounds, initial sample size, budget, and true optimal value (illustrative) for each fidelity level.}
\label{tab:ackley}
\renewcommand{\arraystretch}{1}
\begin{adjustbox}{max width=\textheight} 
\begin{tabular}{p{1cm} p{8cm} p{1.2cm} p{1.2cm} p{1.2cm} p{1.5cm}}
\hline
Agent & Formulation & Upper/lower bound & Init. Sample & Budget & True optimal \\
\hline
1 &
$\begin{aligned}
y_0(x) = & -20 \exp\left(-0.2 \sqrt{0.5 \sum x_i^2}\right) \\
& - \exp\left(0.5 \sum \cos(\pi x_i)\right) + 20 + e
\end{aligned}$
&
$-5 \le x_i \le 5$
&
5 &
50 &
0.000\\
2 &
$\begin{aligned}
y_1(x) = & -20 \exp\left(-0.2 \sqrt{0.5 \sum (x_i+0.2)^2}\right) \\
& - \exp\left(0.5 \sum \cos(1.1\pi (x_i+0.2))\right) + 20 + e + 2.5
\end{aligned}$
&
&
5 &
50 &
2.500 \\
3 &
$\begin{aligned}
y_2(x) = & -20 \exp\left(-0.2 \sqrt{0.5 \sum (0.8(x_i-0.3))^2}\right) \\
& - \exp\left(0.5 \sum \cos(0.9\pi \cdot 0.8(x_i-0.3))\right) + 20 + e + 1.0
\end{aligned}$
&
&
5 &
50/25 &
1.000\\
4 &
$\begin{aligned}
y_3(x) = & -20 \exp\left(-0.2 \sqrt{(x_1+0.4)^2}\right) \\
& - \exp\left(\cos(\pi (x_1+0.4))\right) + 20 + e + 3.0
\end{aligned}$
&
&
5 &
50/25 &
3.000 \\
5 &
$\begin{aligned}
y_4(x) = & -20 \exp\left(-0.2 \sqrt{0.5 \sum (x_i-0.5)^2}\right) \\
& - 1.5 \exp\left(0.5 \sum \cos(\pi (x_i-0.5))\right) + 20 + e + 1.0
\end{aligned}$
&
&
5 &
50 &
-0.359 \\
6 &
$\begin{aligned}
y_5(x) = & 1.1 \Bigl[-20 \exp\left(-0.2 \sqrt{0.5 \sum (x_i-0.1)^2}\right) \\
& - \exp\left(0.5 \sum \cos(\pi (x_i-0.1))\right) + 20 + e \Bigr] \\
& + 4.0
\end{aligned}$
&
&
5 &
50/25 &
3.978 \\
\hline
\end{tabular}
\end{adjustbox}
\end{sidewaystable*}

\begin{sidewaystable*}[h!]
\centering
\caption{Borehole function with five agents. Each agent is assigned a modified variant of the borehole function. Variable bounds, shareable inputs, initial sample size, budget, and true optimal value are listed.}
\label{tab:borehole}
\renewcommand{\arraystretch}{1}
\begin{adjustbox}{max width=\textheight}
\begin{tabular}{p{1.2cm} p{6.0cm} p{3.0cm} p{1.2cm} p{1.2cm} p{1.2cm} p{1.8cm}}
\hline
Agent & Formulation & Variable Bounds & Shareable Inputs & Init. Sample & Budget & True Optimal \\
\hline
1 &
$y_1(\mathbf{x}) = \frac{2\pi T_u (H_u - H_l)}{\ln(r/r_w) \left(1 + \frac{2LT_u}{r_w^2K_w \ln(r/r_w)} + \frac{T_u}{T_l}\right)}$ &
\multirow{5}{3.0cm}{
\begin{minipage}[t]{3.0cm}
\vspace{0.25em}
$0.05 \leq r_w \leq 0.15$ \\
$100 \leq r \leq 10000$ \\
$100 \leq T_u \leq 1000$ \\
$990 \leq H_u \leq 1110$ \\
$10 \leq T_l \leq 500$ \\
$700 \leq H_l \leq 820$ \\
$1000 \leq L \leq 2000$ \\
$6000 \leq K_w \leq 12000$ \\
\vspace{0.25em}
\end{minipage}
} &
\multirow{5}{2.8cm}{
\begin{minipage}[t]{2.8cm}
\vspace{0.25em}
$r_w$ \\
$T_u$ \\
$H_u$ \\
$T_l$ \\
$H_l$ \\
\vspace{0.25em}
\end{minipage}
} & 8 & 50 & 3.985 \\
2 &
$y_2(\mathbf{x}) = \frac{2\pi T_u (H_u - 0.8 H_l)}{\ln(r/r_w) \left(1 + \frac{LT_u}{r_w^2K_w \ln(r/r_w)} + \frac{T_u}{T_l}\right)}$ &
& & 8 & 25 & 15.582 \\
3 &
$y_3(\mathbf{x}) = \frac{2\pi T_u (H_u - H_l)}{\ln(r/r_w) \left(1 + \frac{8LT_u}{r_w^2K_w \ln(r/r_w)} + 0.75\frac{T_u}{T_l}\right)}$ &
& & 8 & 25 & 1.000 \\
4 &
$y_4(\mathbf{x}) = \frac{2\pi T_u (1.09 H_u - H_l)}{\ln(4r/r_w) \left(1 + \frac{3LT_u}{r_w^2K_w \ln(r/r_w)} + \frac{T_u}{T_l}\right)}$ &
& & 8 & 50 & 3.434 \\
5 &
$y_5(\mathbf{x}) = \frac{2\pi T_u (1.05 H_u - H_l)}{\ln(2r/r_w) \left(1 + \frac{3LT_u}{r_w^2K_w \ln(r/r_w)} + \frac{T_u}{T_l}\right)}$ &
& & 8 & 25 & 3.153 \\
\hline
\end{tabular}
\end{adjustbox}
\end{sidewaystable*}

\begin{sidewaystable*}[h!]
\small
\centering
\caption{Wing weight function with four agents. Each agent is assigned a modified variant of the wing weight function. Variable bounds, shareable inputs, initial sample size, budget, and true optimal value are listed.}
\label{tab:wingweight}
\renewcommand{\arraystretch}{1}
\begin{adjustbox}{max width=\textheight}
\begin{tabular}{p{0.8cm} p{6.5cm} p{2.7cm} p{1.8cm} p{0.9cm} p{0.9cm} p{1.4cm}}
\hline
Agent & Formulation & Variable Bounds & Shareable Inputs & Init. Sample & Budget & True Optimal \\
\hline
1 &
$y_1(\mathbf{x}) = 0.036\, s_w^{0.758} w_{fw}^{0.0035} \left(\frac{A}{\cos^2(\Lambda)}\right)^{0.6} \newline
q^{0.006} \lambda^{0.04} \left(\frac{100\, t_c}{\cos(\Lambda)}\right)^{-0.3} 
\left((N_z W_{dg})^{0.49} + s_w w_p\right)$ &
\multirow{4}{3.2cm}{
\begin{minipage}[t]{3.2cm}
\vspace{0.25em}
$150 \leq s_w \leq 200$ \\
$220 \leq w_{fw} \leq 300$ \\
$6 \leq A \leq 10$ \\
$-10 \leq \Lambda \leq 10$ \\
$16 \leq q \leq 45$ \\
$0.5 \leq \lambda \leq 1$ \\
$0.08 \leq t_c \leq 0.18$ \\
$2.5 \leq N_z \leq 6$ \\
$1700 \leq W_{dg} \leq 2500$ \\
$0.025 \leq w_p \leq 0.08$ \\
\vspace{0.25em}
\end{minipage}
} &
\multirow{4}{2.8cm}{
\begin{minipage}[t]{2.8cm}
\vspace{0.25em}
$s_w$ \\
$w_{fw}$ \\
$A$ \\
$q$ \\
$W_{dg}$ \\
\vspace{0.25em}
\end{minipage}
} & 5 & 30 & 123.25 \\
2 &
$y_2(\mathbf{x}) = 0.036\, s_w^{0.758} w_{fw}^{0.0035} \left(\frac{A}{\cos^2(\Lambda)}\right)^{0.6} \newline
q^{0.006} \lambda^{0.04} \left(\frac{100\, t_c}{\cos(\Lambda)}\right)^{-0.3} 
\left((N_z W_{dg})^{0.49} + w_p\right)$ &
& & 5 & 10 & 119.53 \\
3 &
$y_3(\mathbf{x}) = 0.036\, s_w^{0.758} w_{fw}^{0.0035} \left(\frac{A}{\cos^2(\Lambda)}\right)^{0.6} \newline
q^{0.005} \lambda^{0.04} \left(\frac{100\, t_c}{\cos(\Lambda)}\right)^{-0.3} 
\left((N_z W_{dg})^{0.49} + w_p\right)$ &
& & 5 & 20 & 131.65 \\
4 &
$y_4(\mathbf{x}) = 0.036\, s_w^{0.9} w_{fw}^{0.0035} \left(\frac{A}{\cos^2(\Lambda)}\right)^{0.6} \newline
q^{0.005} \lambda^{0.04} \left(\frac{100\, t_c}{\cos(\Lambda)}\right)^{-0.3} 
(N_z W_{dg})^{0.49}$ &
& & 5 & 20 & 268.13 \\
\hline
\end{tabular}
\end{adjustbox}
\end{sidewaystable*}



\printbibliography

\end{document}